\documentclass[letterpaper]{article} 
\usepackage[preprint]{aaai2027}  
\usepackage[hyphens]{url}  
\usepackage{graphicx} 
\usepackage{natbib}  
\usepackage{caption} 

\usepackage{booktabs}
\usepackage{amsmath,amssymb}
\usepackage{multirow}
\usepackage{xcolor}
\usepackage{xcolor}
\usepackage{upquote}   
\usepackage{listings}

\lstdefinestyle{code}{%
  basicstyle=\ttfamily\scriptsize, breaklines=true, breakatwhitespace=false,
  breakindent=0pt, breakautoindent=false,
  postbreak={\mbox{\textcolor{gray}{$\hookrightarrow$}\space}},
  columns=fullflexible, keepspaces=true, showstringspaces=false,
  upquote=true, xleftmargin=2pt, aboveskip=4pt, belowskip=4pt,
}
\lstdefinestyle{out}{style=code, basicstyle=\ttfamily\tiny, frame=single, framesep=2pt}
\newcommand{\method}{CASD}
\newcommand{\best}[1]{\textbf{#1}}
\newcommand{\sd}[1]{{\scriptsize$\pm$#1}}

\title{Coding Agents are Strong Prompt Optimizers}
\author{%
  \large
  \begin{tabular}{@{}c@{\hspace{2.5em}}c@{\hspace{2.5em}}c@{}}
    Agamdeep Singh  & Srishti Gautam & Priyanshu Gupta \\[0.4em]
    Nikita Mehrotra & Tanmay Bakshi  & Sumit Gulwani
  \end{tabular}%
}
\affiliations{%
  \small Microsoft \\[0.3em]
  \footnotesize\texttt{\{t-agasingh, srgautam, priyansgupta,
                        nmehrotra, t-tbakshi, sumitg\}@microsoft.com}%
}

\begin{document}
\maketitle

\begin{abstract}
Search-based prompt optimizers improve prompts through iterative search: they propose edits, execute fresh rollouts, score the resulting trajectories, and retain only edits that improve a validation metric. We show that this optimization loop is unnecessary. Given only a static corpus of agent trajectories, an off-the-shelf coding agent can directly synthesize an optimized prompt, requiring neither environment access nor validation data. We call this approach \emph{Coding-Agent Skill Distillation} (\method{}). The key insight is reflection scope. Rather than reasoning over a small batch of trajectories at each optimization step, the coding agent writes and executes analysis code to compute corpus-wide statistics, identifies systematic failure modes, inspects representative episodes, and distills the resulting insights into behavioral rules. Across four agentic benchmarks (ALFWorld, $\tau^2$-bench retail and telecom, and SpreadsheetBench-Verified), under matched data access, a single \method{} pass outperforms GEPA, a state-of-the-art reflective prompt optimizer, on three of four benchmarks and outperforms validation-gated reflective search (SkillOpt) on all four, improving the unoptimized baseline by 16.6 percentage points on average versus 10.9 for GEPA and 5.3 for SkillOpt. Because \method{} performs a single offline analysis pass rather than iterative search, producing an optimized prompt costs approximately \$1.60---over $22\times$ cheaper than validation-gated search. Even when competing methods are granted additional validation data and unrestricted environment access, \method{} remains ahead on two of four benchmarks. These results suggest that corpus-scale statistical reflection is a viable alternative to iterative search for prompt optimization.
\end{abstract}

\begin{figure*}[t]
\centering
\includegraphics[width=0.90\textwidth]{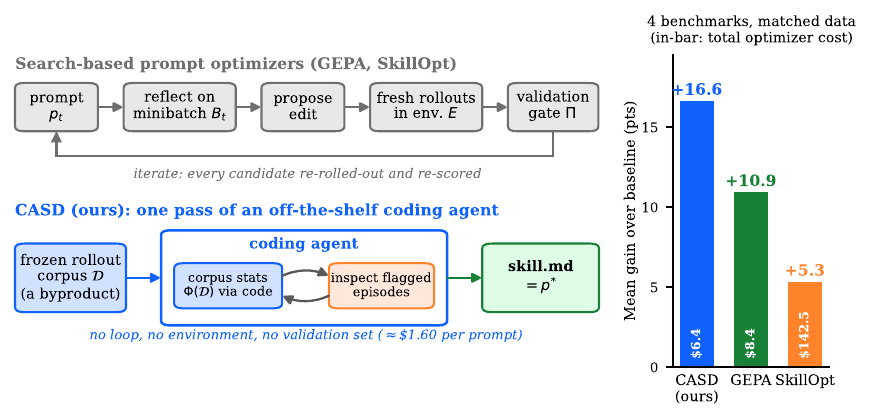}
\caption{\textbf{A coding agent as a prompt optimizer.} \emph{Left:} Search-based optimizers (GEPA, SkillOpt) iteratively refine prompts using environment rollouts and validation feedback. In contrast, \method{} performs a single offline pass over a frozen rollout corpus to directly produce the optimized skill file, without iterative search or additional environment interaction. \emph{Right:} Across four agentic benchmarks under matched data access, \method{} achieves the largest average improvement while requiring substantially lower optimization cost.}
\label{fig:teaser}
\end{figure*}

\section{Introduction}


Large language model (LLM) agents are highly sensitive to their system prompts, motivating a growing line of work on automatic prompt optimization. Search-based prompt optimizers treat the prompt as a learnable artifact by iteratively proposing prompt edits, evaluating them through fresh environment rollouts, and retaining only those that improve a validation objective~\citep{zhou2023ape,yang2024opro,pryzant2023protegi,khattab2024dspy,opsahlong2024mipro}. Recent methods such as GEPA \citep{agrawal2025gepa} strengthen this search process with natural-language reflection, using language models to diagnose failures from sampled trajectories before proposing revised prompts. Despite these advances, the underlying optimization paradigm remains the same: iterative search driven by repeated rollouts and validation.

This search-based recipe carries two structural limitations. First, every prompt revision must be validated through fresh environment interaction, keeping the environment, user simulator, and evaluation metric in the \emph{optimization loop} while making cost grow with the number of candidate edits. More fundamentally, each prompt revision is informed by only a small sample of trajectories, limiting the optimizer's ability to identify behavioral patterns that emerge only at corpus scale. For example, no single reflection step can reliably discover that a tool was invoked 284 times but duplicated 123 times, or that an entire task category failed in 24 of 24 attempts.


We show that the optimization loop can be replaced by a single offline analysis pass over a static corpus of agent trajectories. Our approach, \emph{Coding-Agent Skill Distillation} (\method{}; Figure~\ref{fig:teaser}), uses an unmodified, off-the-shelf coding agent to analyze the corpus and directly synthesize an optimized prompt. The recipe is deliberately simple: provide the coding agent with the rollout corpus and a short natural-language instruction, then use the generated skill file as the optimized system prompt. No optimization loop, validation gate, environment interaction, or held-out validation set is required; prompt optimization becomes a single offline pass costing about \$1.60.


What replaces iterative search is \emph{corpus-scale reflection}. Rather than reasoning over a small sample of trajectories, the coding agent writes and executes analysis code to compute corpus-wide statistics—per-category pass rates, tool-call histograms, duplicate-call counts, and argument-hallucination frequencies—before drilling into the episodes those statistics identify as most informative. Offloading the counting to an interpreter is the same move that makes program-aided prompting exact where free-form reasoning is not \citep{gao2023pal,chen2023pot}, applied here to the optimizer rather than to the task solver. The resulting prompts are grounded in measured evidence rather than anecdotal observations from a handful of trajectories.


Across four agentic benchmarks, a single \method{} pass outperforms GEPA, a state-of-the-art reflective prompt optimizer, on three of four benchmarks and validation-gated reflective search (SkillOpt) on all four when every optimizer is restricted to the same static rollout corpus. Even when the search baselines are granted additional validation data and unrestricted environment access that \method{} never uses, \method{} remains competitive, outperforming them on two of four benchmarks. Our contributions are as follows:


\begin{itemize}
\item We characterize \emph{reflection scope} as a fundamental design dimension for prompt optimization, showing that expanding reflection from sampled trajectories to corpus-scale analysis can replace iterative validation-gated search.

\item We present \emph{Coding-Agent Skill Distillation} (\method{}), a prompt optimization framework that replaces iterative search with a single offline corpus-analysis pass using an unmodified off-the-shelf coding agent, eliminating validation loops and additional environment interaction during optimization.

\item We evaluate \method{} across four agentic benchmarks, showing that a single offline optimization pass consistently matches or outperforms state-of-the-art search-based prompt optimizers while substantially reducing optimization cost.
\end{itemize}

\section{Related Work}

\textbf{Prompt optimization as search.} Discrete prompt search begins with gradient-guided token search \citep{shin2020autoprompt} and moves to LM-driven proposal: APE \citep{zhou2023ape} and OPRO \citep{yang2024opro} sample candidate instructions and keep the best under a task metric; ProTeGi \citep{pryzant2023protegi} follows natural-language ``textual gradients,'' generalized by TextGrad \citep{yuksekgonul2024textgrad} and Trace \citep{cheng2024trace} into backpropagation-like updates over compound systems; DSPy/MIPRO \citep{khattab2024dspy,opsahlong2024mipro} jointly tune instructions and demonstrations under a validation score. A parallel evolutionary line---Promptbreeder \citep{fernando2023promptbreeder}, EvoPrompt \citep{guo2024evoprompt}---mutates and recombines a population, while PromptAgent \citep{wang2024promptagent} plans edits with MCTS. GEPA \citep{agrawal2025gepa} is the strongest reflective variant: it evolves a population of prompts, mutating them with LM reflection over sampled trajectories and selecting on a Pareto front over a validation set. What unites the family is a scoring gate: every candidate must be re-executed and re-scored, so all of them need fresh rollouts per candidate. \method{} needs none.

\textbf{Learning from experience without weight updates.} Reflexion \citep{shinn2023reflexion} and Self-Refine \citep{madaan2023selfrefine} convert per-episode failures into verbal feedback for the next attempt of the \emph{same} task instance. Voyager \citep{wang2023voyager} and ExpeL \citep{zhao2024expel} accumulate reusable skills or insights across episodes; Agent Workflow Memory \citep{wang2024awm} induces reusable workflows from past trajectories; memory-stream architectures \citep{park2023generativeagents} retrieve and reflect over stored observations. A recent wave keeps this loop online at test time---Dynamic Cheatsheet \citep{suzgun2025cheatsheet}, ReasoningBank \citep{ouyang2025reasoningbank}, and ACE \citep{zhang2025ace} curate an evolving context from the agent's own execution feedback. All of these systems reflect trajectory-by-trajectory, with the LM's context window as the bottleneck and with long-context recall degrading as the digest grows \citep{liu2024lostmiddle}. \method{} differs in \emph{how} the corpus is digested: by executable analysis code whose outputs (exact counts over all episodes) then direct targeted reading, so corpus size enters through the interpreter rather than through the context window.

\textbf{Coding agents.} Tool-using coding agents \citep{yang2024sweagent,wang2025openhands,anthropic2025claudecode} interleave code execution, file inspection, and editing to resolve software tasks, and are now strong enough on real repository-level benchmarks \citep{jimenez2024swebench} to be treated as general-purpose analysts rather than code generators. We repurpose one, unmodified, as an optimizer: the ``program'' it edits is a prompt, and the ``test suite'' it consults is a corpus of frozen rollouts.

\textbf{Offline improvement and conservatism.} Improving a policy from a fixed dataset without further interaction is the offline RL setting \citep{levine2020offline}, where the central difficulty is over-estimation on out-of-distribution actions; BCQ \citep{fujimoto2019bcq} and CQL \citep{kumar2020cql} address it by constraining the improved policy to the data support, and one-step methods \citep{brandfonbrener2021onestep} show that a \emph{single} un-iterated improvement step is often preferable to iterated updates when the dataset is small. \method{} is the textual analogue: a single, support-constrained improvement step over a frozen corpus, with no off-policy evaluation to gate it.

\section{Method: Coding-Agent Skill Distillation}


\subsection{Problem Setup}

Let $\pi_\theta(p)$ denote the target agent, where the model parameters
$\theta$ are fixed and only the system prompt $p$ is optimized. Executing
the initial prompt $p_0$ over a training set produces a rollout corpus
\[
\mathcal{D}=\{\tau_i\}_{i=1}^{N},
\]
where each trajectory $\tau_i$ records the complete agent execution,
including messages, tool calls, tool outputs, rewards, and execution
metadata.
The goal of offline prompt optimization is to construct an improved prompt
$p^\star$ from the fixed rollout corpus alone,
\[
p^\star=\mathrm{CASD}(\mathcal{D},p_0),
\]
without collecting additional trajectories or interacting with the
environment. The remainder of this section describes how CASD implements
this offline distillation process.

\subsection{The Distillation Pass}

To instantiate \method{}, we invoke an unmodified off-the-shelf coding
agent with access to the rollout corpus $\mathcal{D}$ and the initial
prompt $p_0$. The agent receives a single high-level natural-language
instruction to analyze the rollout corpus directly and produce an
improved system prompt. Importantly, we do not prescribe an analysis
pipeline, optimization procedure, evaluation metric, or intermediate
representation. Instead, the coding agent determines its own analysis process for extracting useful behavioral rules from the rollout corpus.

\begin{quote}\small
\emph{``There is a results file here containing $N$ agent rollout trajectories
for [task family]. Analyze it directly---no other inputs, no precomputed
summaries---and distill a skill markdown file capturing the behavioral
rules that would make a future agent instance more accurate and more
token/step-efficient on this task family. Use your own judgment fully on
methodology. Write the skill file into this directory.''}
\end{quote}

The distillation process terminates when the agent writes the skill
markdown file, which we use directly as the optimized prompt $p^\star$.
The next subsection characterizes the analysis strategy that emerges from this unconstrained instruction.

\begin{figure*}[t]
\centering
\includegraphics[width=0.90\textwidth]{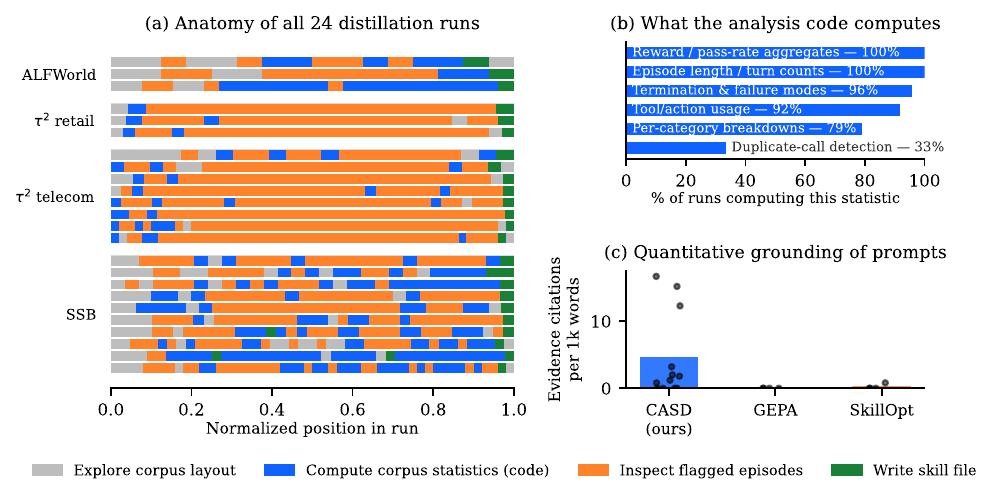}
\caption{Inside the distiller. \textbf{(a)} Tool traces across 24 distillation runs show a characteristic workflow: exploration is front-loaded, skill writing is terminal, and the intermediate steps alternate between corpus-statistics analysis and inspection of flagged episodes. \textbf{(b)} Corpus statistics most frequently computed by the executed analysis code. \textbf{(c)} The resulting prompts consistently cite quantitative evidence (fractions, percentages, and measured counts), reflecting the corpus-driven reasoning process.}
\label{fig:agent}
\end{figure*}

\subsection{Inside the distillation process}
\label{sec:behavior}

Given only the high-level instruction described above, the coding agent
exhibits consistent analysis behavior across independent
distillation runs. To characterize this behavior, we manually classify
every tool call in the 24 logged distillation runs (816 total tool calls)
into four semantic categories: \emph{exploring} the corpus layout (filesystem navigation; 94 calls), \emph{computing corpus statistics} (executing aggregation code over all episodes; 184), \emph{inspecting episodes} (reading individual trajectories, either through code or file inspection; 510), and \emph{writing} the final skill file (28). Figure~\ref{fig:agent} summarizes every run. Three consistent analysis patterns emerge:

\begin{enumerate}

\item \textbf{Exploration precedes synthesis.}
Exploration is front-loaded (median first occurrence at position 0.0 of the run; 34\% of tool calls occur in the first fifth of the run and only 5\% thereafter), while skill writing is consistently terminal (median position 1.0). Individual runs contain 16--51 tool calls (mean 34.0) and produce a 5--8\,KB skill file.

\item \textbf{Statistics-guided investigation.}
Rather than reading trajectories sequentially, the agent alternates between corpus-level statistical analysis and targeted inspection of individual episodes. It analyzes categories with poor pass rates, unusually long trajectories, repeated tool invocations, or early termination, then reads the corresponding trajectories to verify hypotheses before deciding what to investigate next. Runs switch between these two modes a median of five times (up to 16), suggesting an iterative analysis strategy despite the prompt providing no prescribed analysis procedure.

The executed analysis code consistently computes corpus-level statistics rather than simple summaries. Every run measures rewards/pass rates and episode lengths, 96\% quantify termination and failure modes, 92\% analyze tool usage, 79\% break performance down by task category, and one third explicitly search for duplicated tool invocations (Figure~\ref{fig:agent}b).

\item \textbf{Evidence-grounded rule synthesis.}
The resulting skills directly reference the measured behaviors they seek to correct (e.g., ``\emph{16/50 episodes fabricated an identity-lookup argument}'' or ``\emph{\texttt{get\_details\_by\_id} was called 284 times, 123 of them exact duplicates}''). This grounding is reflected in the generated artifacts themselves: the 12 \method{} skills contain 54 quantitative evidence citations (fractions, percentages, and measured counts; 4.6 per 1k words, Figure~\ref{fig:agent}c), compared with none across the four GEPA prompts and only one across the four SkillOpt prompts.

\end{enumerate}

Taken together, these observations show that the coding agent follows a
consistent workflow: it first computes corpus-level statistics, then
uses those statistics to guide targeted inspection of representative
episodes, and finally synthesizes evidence-backed behavioral rules into
an improved prompt. The next section analyzes why this offline
optimization regime behaves differently from iterative search-based
optimizers.

\section{Prompt Optimization Through Corpus-Scale Reflection}
\label{sec:operator}

Although recent prompt optimization algorithms differ operationally, they
all optimize the same objective: improving a prompt from execution
experience to maximize the expected reward of the target agent. Their
primary distinction therefore lies not in \emph{what} they optimize, but
in \emph{how} prompt improvements are inferred. Search-based optimizers
repeatedly estimate prompt updates from sampled trajectories and validate
them through additional rollouts, whereas \method{} performs a single
offline inference from corpus-level statistics computed over a fixed
rollout corpus. We first present a unified formulation of prompt
optimization before analyzing the statistical trade-offs induced by these
different optimization regimes.

\subsection{A Unified View of Prompt Optimization}

Let
\[
J(p)=
\mathbb{E}_{x\sim P_X}
\!\left[
R(\pi_\theta(p),x)
\right]
\]
denote the expected task reward of the target agent under system prompt
$p$. Prompt optimization seeks an improved prompt that maximizes $J(p)$
using execution experience. A broad class of prompt optimization methods
can be written in the common form

\begin{equation}
p_{t+1}
=
\Pi
\Big[
p_t
\oplus
\rho(p_t,\phi_t)
\Big],
\label{eq:template}
\end{equation}

where $\rho$ proposes a prompt modification from execution feedback
$\phi_t$, $\oplus$ applies the modification, and $\Pi$ determines
whether the updated prompt is accepted.

Search-based methods instantiate this framework using sampled rollout
batches together with validation estimates,

\begin{equation}
\begin{aligned}
\phi^{\mathrm{search}}_t
&=
\Big(
\{(\tau,r)\}_{x\in B_t},
\widehat r_{\mathrm{val}}(p)
\Big),\\
\widehat r_{\mathrm{val}}(p)
&=
\frac{1}{|D_{\mathrm{val}}|}
\sum_{x\in D_{\mathrm{val}}}
R(\pi_\theta(p),x),
\end{aligned}
\label{eq:phi-search}
\end{equation}

where $B_t$ denotes the sampled rollout batch, $D_{\mathrm{val}}$ the
validation set, and both the execution feedback and validation score are
Monte Carlo estimates computed from sampled trajectories.

In contrast, \method{} instantiates the same framework using a fixed
rollout corpus $\mathcal D$,

\begin{equation}
\phi^{\method{}}
=
\Big(
\Phi(\mathcal D),
\{\tau_i\}_{i\in\mathcal I(\Phi)}
\Big),
\label{eq:phi-casd}
\end{equation}

where $\Phi(\mathcal D)$ denotes corpus statistics computed over the
rollout corpus, and $\mathcal I(\Phi)$ denotes the index set of
representative trajectories selected for detailed inspection. The next
subsection analyzes the statistical consequences of these two feedback
representations.
\subsection{Bias--Variance Analysis of Prompt Optimization}

The feedback representations in
Equations~(\ref{eq:phi-search}) and~(\ref{eq:phi-casd}) induce different
statistical properties for prompt optimization. For search-based methods,
the validation score $\widehat r_{\mathrm{val}}(p)$ is a Monte Carlo
estimate whose variance decreases with the number of validation
episodes,
\[
\operatorname{Var}
\!\left[
\widehat r_{\mathrm{val}}(p)
\right]
=
\frac{\sigma^2}{|D_{\mathrm{val}}|},
\]
where $\sigma^2$ denotes the per-episode reward variance. Consequently,
prompt updates are inferred from noisy feedback whose reliability
improves only through additional rollout evaluations.

By contrast, the corpus statistics $\Phi(\mathcal D)$ are deterministic
functions of the observed rollout corpus. Once the rollout corpus has
been collected, they eliminate the minibatch sampling variance
associated with repeatedly estimating feedback from sampled trajectory
subsets.

The resulting optimization procedures therefore occupy different points
on the bias--variance trade-off. Search-based optimization relies on
Monte Carlo feedback with decreasing variance as additional rollouts are
collected. In contrast, \method{} reduces estimator variance by
aggregating evidence over the entire rollout corpus, at the cost of
introducing bias whenever the rollout corpus fails to capture important
behaviors. The next subsection analyzes the practical implications of
this trade-off.
\subsection{Operating Regimes}

The preceding analysis suggests that the effectiveness of prompt
optimization depends on the available optimization budget. Under
limited rollout budgets, reducing estimator variance is often more
valuable than eliminating asymptotic bias. In this regime, \method{}
can simultaneously improve optimization effectiveness while operating
at substantially lower optimization cost by replacing iterative search
with a single offline distillation step.

As additional rollout data and environment interaction become
available, the variance of search-based optimization decreases while the
bias associated with a fixed rollout corpus remains unchanged.
Consequently, the relative advantage of offline distillation is expected
to diminish, and iterative search becomes increasingly attractive as
its lower asymptotic bias begins to dominate.

These observations suggest that offline corpus-scale reflection and
iterative search occupy complementary operating regimes rather than
optimizing different objectives. Offline distillation is particularly
well suited to budget-constrained optimization, whereas iterative search
is expected to benefit more from abundant interaction budgets.
\section{Experimental Setup}

\textbf{Benchmarks.} (i)~\textbf{ALFWorld} \citep{shridhar2021alfworld}: embodied household tasks, ReAct-style scaffold \citep{yao2023react}, win rate on 50 held-out games. (ii/iii)~\textbf{$\tau^2$-bench} retail and telecom \citep{barres2025tau2}, the dual-control successor to $\tau$-bench \citep{yao2024taubench}: tool-using customer-service agents against an LM user simulator, pass$@$1 on 40 held-out tasks. (iv)~\textbf{SpreadsheetBench-Verified (SSB)}, derived from SpreadsheetBench \citep{ma2024spreadsheetbench}: spreadsheet manipulation, modified accuracy on 50 held-out items.

\textbf{Models.} Target agent (and user simulator, where applicable): GPT-5.4-mini with reasoning disabled (\emph{no-think}) throughout. Optimizer/reflection LM for all methods: Claude Sonnet~5. Test accuracy is the mean over 3 seeds; parenthesized values are sample SD ($n{-}1$). At these evaluation-set sizes the binomial standard error alone is $\approx$7 points per seed, so we report seed-level dispersion throughout rather than single point estimates \citep{miller2024errorbars}.

\textbf{Methods.} \emph{Baseline}: no skill, cost \$0. \emph{\method{} (ours)}: one distillation pass per skill over the static pool rollouts; we synthesize 3 skills per benchmark and report their mean (so our SD measures skill-to-skill variance; baselines' SD measures seed-to-seed variance of one skill). \emph{GEPA} \citep{agrawal2025gepa}: evolutionary reflective prompt optimization. \emph{SkillOpt}: reflective search with a validation-selection gate.

\textbf{Data regimes.} In the \emph{limited-data} regime---our headline comparison---every optimizer sees only the same fixed pool (retail 35, telecom 50, SSB 50 tasks; a 50-game ALFWorld pool): GEPA sets its Pareto set equal to train, SkillOpt splits the pool internally. In the \emph{head-to-head} regime, GEPA and SkillOpt additionally receive a separate held-out validation set (and, as always, unlimited environment access for candidate rollouts) that \method{} never uses.


\textbf{Optimizer settings and hyperparameters} For reproducibility: all methods use Claude Sonnet~5 as the
reflection/optimizer LM. GEPA runs with
a budget of 120 metric calls, with its
Pareto set equal to train in the limited-data regime; resuming the telecom run to 240 calls
returned a byte-identical prompt, so we report the 120-call point. SkillOpt runs 3--5 epochs
with minibatch size scaled to the train split ($8$--$40$), an edit budget of 4 per step
(cosine-decayed to 2), and its validation gate enabled; its internal split is $25/10$ on
retail and $35/15$ elsewhere for limited data regime setting. 

\begin{table}[t]
\centering\small
\setlength{\tabcolsep}{3.5pt}
\caption{\textbf{Limited-data regime (headline):} held-out test accuracy (\%), mean over 3 seeds (SD), when all optimizers see only the static rollout pool. $^{\ddagger}$ALFWorld SkillOpt uses its native scaffold (own baseline 58.7\%); not directly comparable to column~2.}
\label{tab:ld}
\begin{tabular}{lcccc}
\toprule
Benchmark & Baseline & \method{} (ours) & SkillOpt & GEPA \\
\midrule
ALFWorld & 56.7\sd{1.2} & \best{83.3}\sd{2.3} & 68.0\sd{2.0}$^{\ddagger}$ & 74.0\sd{2.0} \\
$\tau^2$ retail & 32.5\sd{2.5} & \best{40.0}\sd{2.5} & 38.3\sd{10.1} & 39.2\sd{7.2} \\
$\tau^2$ telecom & 19.2\sd{5.2} & \best{39.2}\sd{8.0} & 25.8\sd{3.8} & 17.5\sd{4.3} \\
SSB-Verified & 39.3\sd{3.1} & 51.3\sd{3.1} & 38.7\sd{2.3} & \best{60.7}\sd{3.1} \\
\midrule
Mean gain vs.\ base & --- & \best{+16.6} & +5.3 & +10.9 \\
\bottomrule
\end{tabular}
\end{table}

\begin{table}[t]
\centering\small
\setlength{\tabcolsep}{3.5pt}
\caption{\textbf{Head-to-head regime:} GEPA and SkillOpt additionally get a separate held-out validation set and environment access; \method{} is unchanged (it uses neither). ALFWorld cells repeat the limited-data runs (no separate head-to-head run exists).}
\label{tab:hh}
\begin{tabular}{lcccc}
\toprule
Benchmark & Baseline & \method{} (ours) & SkillOpt & GEPA \\
\midrule
ALFWorld & 56.7\sd{1.2} & \best{83.3}\sd{2.3} & 68.0\sd{2.0}$^{\ddagger}$ & 74.0\sd{2.0} \\
$\tau^2$ retail & 32.5\sd{2.5} & 40.0\sd{2.5} & 38.3\sd{10.1} & \best{46.7}\sd{8.0} \\
$\tau^2$ telecom & 19.2\sd{5.2} & 39.2\sd{8.0} & \best{44.2}\sd{5.2} & 15.8\sd{5.2} \\
SSB-Verified & 39.3\sd{3.1} & 51.3\sd{3.1} & 48.0\sd{2.0} & \best{56.7}\sd{1.2} \\
\bottomrule
\end{tabular}
\end{table}

\begin{figure}[t]
\centering
\includegraphics[width=\columnwidth]{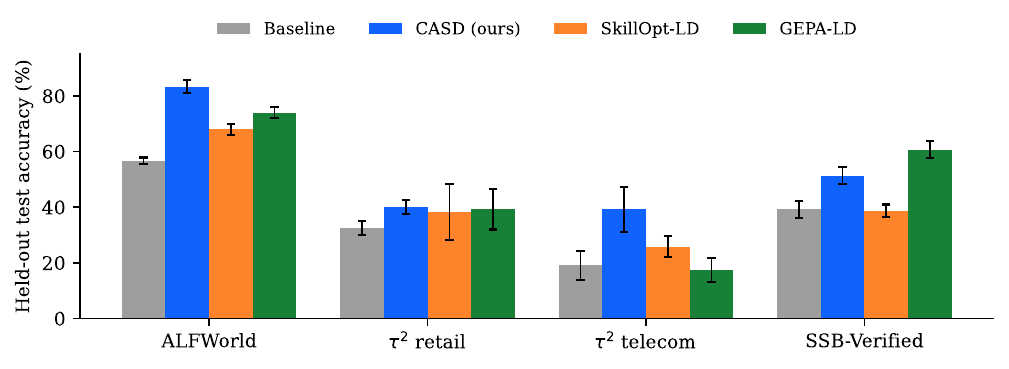}
\caption{Limited-data regime: held-out test accuracy over 3 seeds (bars: mean; whiskers: SD). A single \method{} pass over the static pool beats GEPA on 3/4 benchmarks and SkillOpt on 4/4.}
\label{fig:main}
\end{figure}


\begin{figure*}[t]
\centering
\includegraphics[scale=0.8]{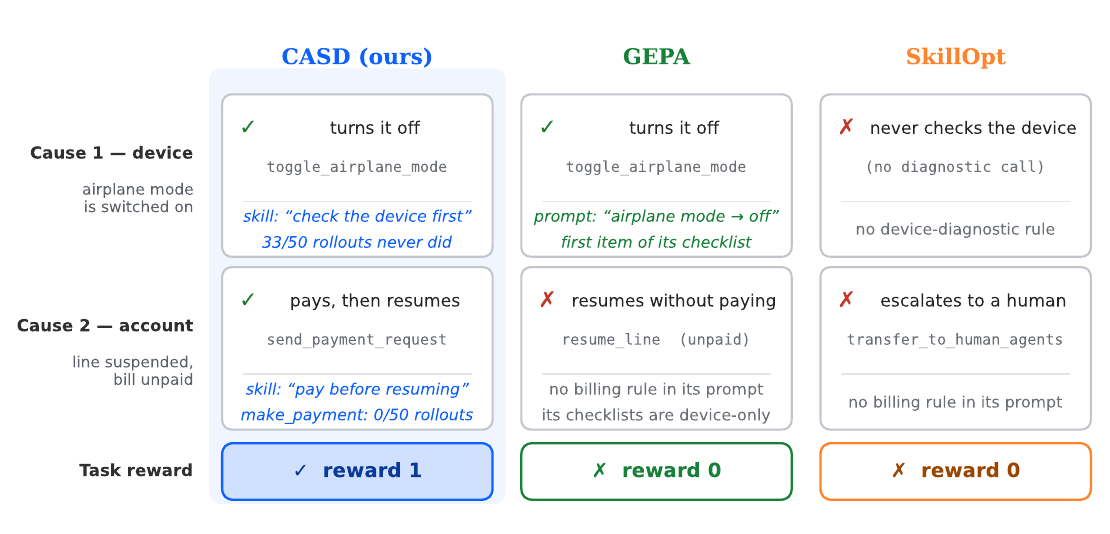}
\caption{\textbf{Corpus-level absences are invisible to minibatch reflection.} A $\tau^2$-telecom ``no service'' ticket requires repairing two independent root causes. Columns show the prompts distilled by the three optimizers from the same 50-rollout pool and the resulting agent behavior. The device issue is evident in individual trajectories and is captured by all methods. In contrast, the billing issue appears only as a corpus-level absence---\texttt{make\_payment} is never invoked in the rollout pool. Only \method{} identifies this missing behavior from corpus-level statistics and distills the required billing rule, whereas GEPA and SkillOpt omit it and fail the task.}
\label{fig:case}
\end{figure*}

\section{Results}

\subsection{Matched data access: one pass beats the loops}
Table~\ref{tab:ld} and Figure~\ref{fig:main} give the headline comparison. With every optimizer restricted to the same static pool, \method{} is best on ALFWorld ($83.3$ vs.\ GEPA's $74.0$), retail ($40.0$ vs.\ $39.2$), and telecom ($39.2$ vs.\ $17.5$), and second on SSB ($51.3$ vs.\ GEPA's $60.7$). Averaged over benchmarks, \method{} lifts the baseline by $+16.6$ points, versus $+10.9$ for GEPA and $+5.3$ for SkillOpt. Telecom is the sharpest separation: with no external validation signal, GEPA's evolutionary loop \emph{degrades} below baseline ($17.5$ vs.\ $19.2$)---its edits are selected on the same pool it reflects on, and overfit---while the coding agent's statistics-first analysis of the identical data yields $+20.0$ points.

\subsection{Head-to-head: search buys back some ground, at a price}
When GEPA and SkillOpt are granted an extra validation set and unrestricted environment rollouts (Table~\ref{tab:hh}), they improve where validation is informative: GEPA reaches $46.7$ on retail and SkillOpt $44.2$ on telecom. \method{}, which touches neither the environment nor a validation set, still wins ALFWorld outright and remains within noise of the best telecom cell given its skill-to-skill SD. Notably GEPA \emph{still} fails on telecom ($15.8$) even with validation---reflective mutation never finds the domain's core failure modes---whereas SkillOpt only fixes telecom by spending \$74.7 of gated search (Table~\ref{tab:cost}).

\begin{table}[t]
\centering\small
\setlength{\tabcolsep}{3.0pt}
\caption{Optimizer cost to \emph{produce} each prompt (USD; with token-caching on). GEPA/SkillOpt telecom and SSB(SkillOpt) costs are from the head-to-head runs. \method{} needs no environment rollouts at all.}
\label{tab:cost}
\begin{tabular}{lcccc|c}
\toprule
 & ALFWorld & Retail & Telecom & SSB & Total \\
\midrule
\method{} (ours) & 1.6 & 1.6 & 1.56 & 1.6 & \best{6.4} \\
GEPA & 2.1 & 1.1 & 4.5 & 0.74 & 8.4 \\
SkillOpt & 15.9 & 13.9 & 74.7 & 38.0 & 142.5 \\
\bottomrule
\end{tabular}
\end{table}

\subsection{Cost}
Table~\ref{tab:cost} summarizes production cost. A \method{} pass costs $\approx$\$1.60 per skill ---one multi-tool coding-agent session, with no rollout bill because the corpus is a byproduct of rollouts that were already generated. Totaled over the four benchmarks this is $22\times$ cheaper than SkillOpt (\$142.5) and still less than GEPA (\$8.4); unlike both, it also runs where no simulator, grader, or validation split exists. Optimized prompts also shift \emph{test-time} cost: e.g., on telecom, SkillOpt's prompt induces 20.1M prompt tokens over the 3-seed evaluation versus 11.9M for GEPA's, a reminder that verbose optimized prompts are not free to deploy.

\subsection{Why does corpus-scope reflection win?}
Three observations support the reflection-scope explanation. \textbf{(1) The rules are statistical, not anecdotal.} The telecom skill's top rule targets fabricated identity-lookup arguments observed in 16/50 episodes; no single-minibatch reflection reliably surfaces a 32\%-frequency error, and GEPA's telecom prompts never address it. \textbf{(2) Efficiency rules need duplicate counts.} Detecting that 123 of 284 \texttt{get\_details\_by\_id} calls were exact duplicates requires joining tool calls across a whole episode set---a one-liner in pandas, but invisible in a context-window reflection. \textbf{(3) No gate, no gate-overfitting.} Search methods keep an edit only if a small validation batch approves, which both overfits small pools \citep{smith2006optimizerscurse,dwork2015reusable} (GEPA-telecom-LD collapsing to $17.5$) and inflates variance (SkillOpt retail SD $10.1$). \method{}'s single pass has no acceptance step to overfit; its variance across independently produced skills is small (SD $\le 3.1$ on 3 of 4 benchmarks).

Figure~\ref{fig:case} makes the mechanism concrete on a single test episode. The ticket has two independent root causes, and reward is granted only if the target agent repairs both. The device-side cause is the kind of failure a minibatch reflection can see---it is visible in any individual failed trajectory---and all three optimizers encode a rule for it. The account-side cause is not: it appears in the corpus only as an \emph{absence}, a payment call that no rollout ever makes, which is a statement about the whole pool rather than about any one episode it contains. Only \method{} writes a rule for it, and only \method{} solves the episode.

\subsection{Ablation: does the skill recover thinking?}
\label{sec:think}
Our target model runs with reasoning disabled, so a natural reference point is the \emph{same} model with reasoning turned on \citep{wei2022cot,deepseek2025r1}---the canonical way to trade output tokens for accuracy at test time \citep{snell2025testtime}. How much of the accuracy that thinking buys can a distilled prompt recover, and at what token cost? Table~\ref{tab:think} compares three modes of GPT-5.4-mini on all four benchmarks: no-think, no-think with the \method{} skill, and think.\footnote{Think accuracies and per-episode token counts come from an earlier study on the same test splits (except SSB think, which used a different harness and is indicative). Baseline and \method{} accuracies are from the current runs in Table~\ref{tab:ld}.}

\begin{table}[t]
\centering\small
\setlength{\tabcolsep}{3.0pt}
\caption{Recovering the reasoning gap without reasoning tokens (GPT-5.4-mini). ``Rec.''\ = fraction of the no-think$\to$think accuracy gap recovered by the skill; tok = mean output tokens per episode (think $\to$ skill).}
\label{tab:think}
\begin{tabular}{lccccc}
\toprule
Benchmark & no-think & +\method{} & think & Rec. & tok ($\downarrow$) \\
\midrule
ALFWorld & 56.7 & \best{83.3} & 71.3\sd{1.2} & $>$100\% & 3.7k$\to$0.8k \\
$\tau^2$ retail & 32.5 & \best{40.0} & 35.0\sd{6.6} & $>$100\% & 1.6k$\to$0.6k \\
$\tau^2$ telecom & 19.2 & 39.2 & \best{45.0}\sd{2.5} & 78\% & 2.1k$\to$0.6k \\
SSB-Verified & 39.3 & 51.3 & \best{61.3}\sd{1.2} & 55\% & 3.3k$\to$0.8k \\
\bottomrule
\end{tabular}
\end{table}


Two findings. \textbf{(1) Distilled rules substitute for much of test-time reasoning.} On ALFWorld and retail the skill \emph{exceeds} the think mode outright ($83.3$ vs.\ $71.3$; $40.0$ vs.\ $35.0$); on telecom and SSB it recovers $78\%$ and $55\%$ of the no-think$\to$think gap. It does so while emitting \emph{zero} reasoning tokens: per-episode output stays at or below the no-think budget (0.6--0.8k tokens), where thinking costs $2.9$--$4.5\times$ more. The interpretation is that much of what reasoning re-derives episode after episode---which diagnostic to run next, when a lookup argument is unjustified, when not to give up---is \emph{policy-like} and can be crystallized once, offline, into explicit rules. \textbf{(2) The distillation corpus need not contain thinking.} In the earlier study round we also distilled skills from \emph{think} rollouts of the same pools (a contrastive think-vs-no-think corpus) and deployed them on the no-think policy: across benchmarks the two corpus compositions land within a few points of each other with no consistent winner (e.g., ALFWorld $81.3$ think-distilled vs.\ $78.7$ no-think-distilled; retail $45.8$ vs.\ $40.8$; telecom $32.5$ vs.\ $33.3$; SSB reversed, $46.0$ vs.\ $56.0$). Failure-rich no-think rollouts alone carry enough signal; expensive reasoning traces are optional corpus enrichment, needed at most once at corpus-construction time. The residual gaps on telecom and SSB suggest a portion of thinking---presumably instance-specific deduction rather than reusable policy---that no static prompt recovers; closing it is future work.

\section{Limitations}
Our SD for \method{} measures skill-to-skill variance (3 independently distilled skills, one evaluation each) while baselines report seed variance of a single prompt; the quantities are close but not identical. All results use one target model (GPT-5.4-mini no-think) and one coding agent (Claude Sonnet~5/Claude Code); the recipe's sensitivity to distiller capability is untested. Finally, \method{} inherits the corpus: it cannot discover behaviors absent from the logged rollouts and very small or failure-free corpora may leave nothing to distill.

\section{Conclusion}
A stock coding agent, pointed at a directory of frozen rollouts with a one-paragraph instruction, is a strong prompt optimizer: it beats state-of-the-art reflective search under matched data access on 3 of 4 agentic benchmarks, never touches the environment, and costs about \$1.60 per prompt. The key difference is reflection scope---executing analysis code over the entire rollout corpus instead of relying on minibatch reflection and validation gating. As coding agents improve, offline corpus-scale reflection may become the default first step of prompt optimization, with search reserved for the final gains when additional interaction is inexpensive.

\bibliography{references}


\end{document}